\documentclass[10pt, a4paper]{article}
\usepackage{lrec2022} % this is the new LREC2022 Style
\usepackage{multibib}
\newcites{languageresource}{Language Resources}
\usepackage{graphicx}
\usepackage{tabularx}
\usepackage{soul}
\usepackage{booktabs}
\usepackage{titlesec}
\titleformat{\section}{\normalfont\large\bfseries\center}{\thesection.}{1em}{}
\titleformat{\subsection}{\normalfont\SmallTitleFont\bfseries\raggedright}{\thesubsection.}{1em}{}
\titleformat{\subsubsection}{\normalfont\normalsize\bfseries\raggedright}{\thesubsubsection.}{1em}{}
\renewcommand\thesection{\arabic{section}}
\renewcommand\thesubsection{\thesection.\arabic{subsection}}
\renewcommand\thesubsubsection{\thesubsection.\arabic{subsubsection}}
\usepackage{epstopdf}
\usepackage[utf8]{inputenc}

\usepackage{hyperref}
\usepackage{xstring}

\usepackage{multirow, enumitem, xcolor, colortbl, subcaption}

\usepackage{color}

\title{Large-Scale Hate Speech Detection with Cross-Domain Transfer}

\name{Cagri Toraman, Furkan \c{Ş}ahinu\c{ç}, Eyup Halit Yilmaz} 

\address{Aselsan Research Center, Ankara, Turkey \\
         \{ctoraman, fsahinuc, ehyilmaz\}@aselsan.com.tr\\}

\abstract{The performance of hate speech detection models relies on the datasets on which the models are trained. Existing datasets are mostly prepared with a limited number of instances or hate domains that define hate topics. This hinders large-scale analysis and transfer learning with respect to hate domains. In this study, we construct large-scale tweet datasets for hate speech detection in English and a low-resource language, Turkish, consisting of human-labeled 100k tweets per each. Our datasets are designed to have equal number of tweets distributed over five domains. The experimental results supported by statistical tests show that Transformer-based language models outperform conventional bag-of-words and neural models by at least 5\% in English and 10\% in Turkish for large-scale hate speech detection.
The performance is also scalable to different training sizes, such that 98\% of performance in English, and 97\% in Turkish, are recovered when 20\% of training instances are used. We further examine the generalization ability of cross-domain transfer among hate domains. We show that 96\% of the performance of a target domain in average is recovered by other domains for English, and 92\% for Turkish. Gender and religion are more successful to generalize to other domains, while sports fail most.
 \\ 
 \newline 
  }

\begin{document}

\maketitleabstract

\section{Introduction}

With the growth of social media platforms, hate speech towards people who do not share the same identity or community becomes more visible \cite{TwitterReport:2021}. Consequences of online hate speech can be real-life violence against other people and communities \cite{Byman:2021}. The need to automatically detect hate speech text is thereby urging. 

Existing solutions to detect hate speech mostly rely on supervised scheme, resulting in a strict dependency on the quality and quantity of labeled data. Most of the datasets labeled by experts for hate speech detection are not large in size due to the labor cost \cite{Poletto:2021}, causing a lack of detailed experiments on model generalization and scalability. Indeed, most studies on hate speech detection report high performances on their test sets, while their generalization capabilities to other datasets can be limited \cite{Arango:2019}. Moreover, existing datasets for hate speech detection are very limited for low-resource languages such as Turkic languages \cite{Poletto:2021}. We thereby construct large-scale datasets for hate speech detection in English and Turkish, consisting of human-labeled 100k tweets per each and compare the performance of state-of-the-art models on these large-scale datasets.

Hateful language can be expressed in various topics (we refer to topics as \emph{hate domains}). Hate domains vary depending on the target group. For instance, misogyny is an example of the domain of gender-based hatred. Existing studies mostly do not consider various domains explicitly. They also investigate hate speech in terms of an abstract notion including aggressive language, threats, slurs, and offenses \cite{Poletto:2021}. We consider not only the hateful behavior in the definition of hate speech, but also five most frequently observed domains depending on target group; namely religion, gender, racism, politics, and sports. 

Supervised models trained on a specific learning dataset can fail to generalize their performance on the original evaluation set to other evaluation sets. This phenomenon is studied in zero-shot cross-dataset\footnote{In literature, the phrase ``cross-domain" is mostly used for the transfer between two datasets that are published by different studies but not necessarily in different hate domains. We refer to them as cross-dataset.} \cite{Grondahl:2018,Karan:2018}, cross-lingual \cite{Pamungkas:2019}, and cross-platform \cite{Agrawal:2018} transfer for hate speech detection. However, transfer learning with respect to hate domains and low-resource languages are not well studied due to the lack of large-scale datasets. In this study, with the help of our datasets containing five hate domains, we analyze the generalization capability of hate speech detection in terms of cross-domain transfer among hate domains.
 
The \textbf{contributions} of this study are in three folds. (i) We construct large-scale human-labeled hate speech detection datasets for English and Turkish. (ii) We analyze the performance of various models for large-scale hate speech detection with a special focus on model scalability. (iii) We examine the generalization capability of hate speech detection in terms of zero-shot cross-domain transfer between hate domains.

In the next section, we provide a summary of related work. In Section \ref{section:datasets}, we explain our large-scale datasets\footnote{The paper contains some examples of language which may be offensive to some readers. They do not represent the views of the authors.}. In Section \ref{section:experiments}, we report our experiments. In Section \ref{section:discussion}, we provide discussions on error analysis % existing models, 
and scalability. %, and an ablation study. 
We conclude the study in the last section.

\section{Related Work}
\label{section:related_work}
We summarize related work on the methods, datasets, and transfer learning for hate speech detection.

\subsection{Methods for Hate Speech Detection}
Earlier studies on hate speech detection are based on matching hate keywords using lexicons \cite{Sood:2012}. The disadvantage of such methods is strict dependency on lexicons. Supervised learning with a set of features extracted from a training set is a solution for the dependency issue. Text content is useful to extract bag-of-words features; such as linguistic and syntactical features, n-grams, and Part-of-Speech tags  \cite{Nobata:2016,Waseem:2016,Davidson:2017}. User-based features, including content history, meta-attributes, and user profile can be used to detect hate signals \cite{Waseem:2016,Chatzakou:2017,Unsvaag:2018}. % Structural features of a social network, e.g., centrality and clustering, are studied as well \cite{Chatzakou:2017}. 

To capture word semantics better than bag-of-words; word embeddings, such as GloVe \cite{Pennington:2014}, are utilized to detect abusive and hatred language \cite{Nobata:2016,Mou:2020}. Character and phonetic-level embeddings are also studied for hate speech to resolve the issues related to noisy text of social media \cite{Mou:2020}. Instead of extracting hand-crafted features; deep neural networks, such as CNN \cite{Kim:2014} and LSTM \cite{Hochreiter:1997}, are applied to extract deep features to represent text semantics. Their application outperforms previous ones with lexicons and hand-crafted features \cite{Zimmerman:2018,Cao:2020}.

Recently, Transformer architecture \cite{Vaswani:2017} is studied for hate speech detection. %, as in all other downstream tasks of NLP. 
Transformer employs self-attention for each token over all tokens, targeting to capture a rich contextual representation of whole text. Fine-tuning a Transformer-based model, BERT, \cite{Devlin:2019} for hate speech detection outperforms previous methods \cite{Liu:2019a,Caselli:2021,Mathew:2021}. We examine the large-scale performance of not only BERT, but also various Transformer-based language models, as well as conventional bag-of-words and neural models.

\subsection{Resources for Hate Speech Detection}
A recent survey summarizes the current state of datasets in hate speech detection by listing over 40 datasets, around half of which are tweets, and again around half of which are prepared in English language \cite{Poletto:2021}. Benchmark datasets are also released as a shared task for hate speech detection \cite{Basile:2019,Zampieri:2020}. 

There are efforts to create large-scale human-labeled datasets for hate speech detection. The dataset by \cite{Davidson:2017} has approximately 25k tweets each labeled by three or more annotators for three classes; offensive, hate, and neither. The dataset by \cite{Golbeck:2017} has 35k tweets labeled by at most three annotators per tweet for binary classification (harassing or not). The dataset by \cite{Founta:2018} has 80k tweets each labeled by five annotators for seven classes including offensive and hate. There also exist studies that construct datasets containing hateful content from various sources (e.g.  Facebook and Reddit) in other low-resource languages; such as Arabic \cite{Albadi:2018}, Greek \cite{Pavlopoulos:2017}, Slovene \cite{Fiser:2017}, and Swedish \cite{Fernquist:2019}. However, our datasets differ in terms of the following aspects. We have 100k top-level tweets per two languages, English and Turkish. The datasets have three class labels (hate, offensive, and normal), and five annotators per each tweet. We focus on dataset cleaning, which will be explained in the next section. Lastly, we design to have 20k tweets for each of five hate domains, enabling us to analyze cross-domain transfer.

\subsection{Transfer Learning for Hate Speech Detection}
Generalization of a hate speech detection model trained on a specific dataset to other datasets with the same or similar class labels, i.e., zero-shot cross-dataset transfer, is widely studied \cite{Karan:2018,Swamy:2019,Arango:2019,Pamungkas:2020,Markov:2021a}. Using different datasets in different languages, cross-lingual transfer aims to overcome language dependency in hate speech detection \cite{Pamungkas:2019,Pamungkas:2020,Markov:2021b,Nozza:2021}. There are also efforts to analyze platform-independent hate speech detection, i.e. cross-platform transfer \cite{Agrawal:2018}. In this study, we analyze whether hate speech detection can be generalized across several hate domains, regardless of the target and topic of hate speech. 

\section{Large-Scale Datasets for Hate Speech Detection}
\label{section:datasets}

% \begin{table}[t]
% \small
% \setlength{\tabcolsep}{1.5pt}
% \centering
% \begin{tabular}{ll}
% \hline
% \textbf{Domain} & \textbf{Keywords} \\
% \hline
% \multirow{2}{*}{Religion} & Christianity, Islam, Judaism, Hinduism, Atheist, \\
% & belief, church, mosque, Jewish, Muslim, Bible \\
% \hline
% \multirow{2}{*}{Gender} & LGBTQ, bisexual, female, male, homophobia, \\
% & gay, lesbian, homosexual, bisexual, transgender \\
% \hline
% \multirow{2}{*}{Race} & foreigner, refugee, immigrant, Syrian, African, \\
% & Turk, American, Iranian, Russian, Arab, Greek \\
% \hline
% \multirow{2}{*}{Politics} & democratic party, republican party, government, \\
% & white house, president, Trump, Biden, minister \\
% \hline
% \multirow{2}{*}{Sports} & football, baseball, volleyball, referee, barcelona, \\
% & real madrid, chelsea, new york knicks, coach \\
% \hline
% \end{tabular}
% \caption{Samples from our keyword list. Turkish keywords are mostly translations of English keywords.}
% \label{tab:keywords}
% \end{table}

\begin{table*}[h]
\small
\setlength{\tabcolsep}{8.0pt}
\renewcommand{\arraystretch}{1.0}
\centering
\begin{tabular}{ll}
\hline
\textbf{Domain} & \textbf{Keywords} \\
\hline
Religion & Christianity, Islam, Judaism, Hinduism, Atheist, belief, church, mosque, Jewish, Muslim, Bible \\
\hline
Gender & LGBTQ, bisexual, female, male, homophobia, gay, lesbian, homosexual, bisexual, transgender \\
\hline
Race & foreigner, refugee, immigrant, Syrian, African, Turk, American, Iranian, Russian, Arab, Greek \\
\hline
Politics & democratic party, republican party, government, white house, president, Trump, Biden, minister \\
\hline
Sports & football, baseball, volleyball, referee, barcelona, real madrid, chelsea, new york knicks, coach \\
\hline
\end{tabular}
\caption{Samples from our keyword list. Turkish keywords are mostly translations of English keywords.}
\label{tab:keywords}
\end{table*}

\begin{table*}[ht]
\small
\centering
\renewcommand{\arraystretch}{1.0}
\begin{tabular}{llc}
\hline
\textbf{Domain} & \textbf{Tweet} & \textbf{Label} \\
\hline
Gender & “I can’t live in a world where gay marriage is legal.” Okay, so die. & Hate \\
\hline
% \multirow{4}{*}{Race}
% & Sorry that I said Turks aren't & \multirow{4}{*}{Hate}\\
% & righteous in killing YPG Kurds &  \\ 
% & Syrians My bad bro I hate those &  \\ 
% & stinky Kurds as much as Gypsies & \\
% \hline
% 
% \multirow{3}{*}{Race} & 5 paralik oy uğruna terörist ile kolkola giren hayin ermeni kökenli yavsaklar allah sizinde  & \multirow{3}{*}{Hate} \\
% & ocağiniza ates dusursun (May God set fire to your hearth, treacherous Armenian-origin  & \\
% & sluts who joined arms with terrorists for the sake of worthless votes.) &  \\ 
Race & Türklere iyi geceler, amerikalılar gebersin (Good night to the Turks, death to the Americans) & Hate \\
\hline
Religion & Self proclaim atheist doesn't make you cool kid bitch & Offensive \\
\hline
% \multirow{2}{*}{Sports} & Referee is so fucking blind Clear & \multirow{2}{*}{Offensive} \\
% & foul on Semedo again not given. & \\
\multirow{2}{*}{Sports} & Bundan sonra 6sn kuralını saymayan Hakem de uygulamayan da Şerefsiz oğlu şerefsiztir & \multirow{2}{*}{Offensive} \\
&  (After that, the referee, who does not count and apply the 6-second rule, will be dishonest.) & \\
\hline
Politics & Biden your a lier and a cheat and a old idiot & Offensive \\
\hline 
\end{tabular}
\caption{Tweet examples (not edited) from the dataset. Translation for Turkish tweets are given in parentheses.}
\label{tab:samples}
\end{table*}

\subsection{Dataset Construction}

We used Full-Archive Search provided by Twitter Premium API to retrieve more than 200k tweets; filtered according to language, tweet type, publish time, and contents. We filter English and Turkish tweets published in 2020 and 2021, since old tweets are more likely to be deleted. The datasets\footnote{The datasets include publicly available tweet IDs, in compliance with Twitter's Terms and Conditions, and can be accessed from https://github.com/avaapm/hatespeech} contain only top-level tweets, i.e., not a retweet, reply, or quote. Tweet contents are filtered based on a keyword list determined by the dataset curators. The list contains hashtags and keywords from five topics (i.e., hate domains); religion, gender, racism, politics, and sports. A tweet can only belong to a single topic. Samples from the complete keyword list with corresponding domains are given in Table \ref{tab:keywords}. We design to keep the number of tweets belonging to each hate domain balanced. To this end, slightly more than 20k tweets are retrieved from Twitter for each domain. The exact amount of 20k tweets are then sampled for each domain to satisfy the balance.

For cleaning, we remove near-duplicate tweets by measuring higher than 80\% text similarity among tweets using the Cosine similarity with TF-IDF weighting. We restrict the average number of tweets per user not exceeding 1\% of all tweets to avoid user-dependent modeling \cite{Geva:2019}. We remove tweets shorter than five words excluding hashtags, URLs, and emojis.

\subsection{Dataset Annotation}

Based on the definitions and categorization of hateful speech \cite{Sharma:2018}, we label tweets as containing hate speech if they target, incite violence against, threaten, or call for physical damage for an individual or a group of people because of some identifying trait or characteristic. We label tweets as offensive if they humiliate, taunt, discriminate, or insult an individual or a group of people in any form, including textual. Other tweets are labeled as normal.

Each tweet is annotated by five annotators randomly selected from a set of 20 annotators, 75\% of which are graduate students while the rest are undergraduate students. 65\% of the annotators's gender of birth are female and 35\% are male. Their ages fall within the range of 20-26. 
While annotating a tweet, if consensus is not achieved on ground-truth, a dataset curator outside the initial annotator set determines the label. The curator intervenes in only 8\% of the total tweets (38\% of tweets are labeled with the consensus of five, 26\% with four, and 28\% with three annotators). We provide a list of annotation guidelines to all annotators. The guidelines document includes the rules of annotations; the definitions of hate, offensive, and normal tweets; and the common mistakes observed during annotation. The annotations started on February 15th, and ended on May 10th, 2021 (i.e. a period of 84 days). We measure inter-annotator agreement with Krippendorff's alpha coefficient and get a nominal score of 0.395 for English and 0.417 for Turkish, which are higher than other similar hate speech datasets in the literature (0.38 in binary \cite{Sanguinetti:2018} and 0.153 in multi-class \cite{Ousidhoum:2019}). Sample hateful and offensive tweets from the datasets are given in Table \ref{tab:samples}.

\subsection{Dataset Statistics}
We report main statistics about our datasets in Table \ref{tab:dataset_stats}. Although we follow a similar construction approach for both languages, the number of tweets with hate speech that we can find in English is less than those in Turkish, which might indicate a stronger regularization of English content by Twitter. Normal tweets dominate as expected due to the nature of hate speech and the platform regulations. The statistics of tweet length imply that our task is similar to a short text classification for tweets, where the average number of words is ideal to be 25 to 30 \cite{Sahinuc:2021}.

The domain and class distributions of tweets are given in Table \ref{tab:tweet_distributions}. In English, the number of hateful tweets is close in each domain; however, race has less number of offensive tweets than others. The number of hateful tweets in gender domain is less than those of other domains in Turkish dataset.

\begin{table}
\small
\centering
\begin{tabular}{lrr}
\hline
\textbf{Definition} & \textbf{EN} & \textbf{TR} \\
\hline
Number of tweets & 100,000 & 100,000 \\
Number of offensive tweets & 27,140 & 30,747 \\
Number of hate tweets & 7,325 & 27,593 \\
Number of users & 85,396 & 69,524 \\
First tweet date & 26/02/20 & 17/01/20 \\
Last tweet date & 31/03/21 & 31/03/21 \\
% Average tweets per user & 1.171 & 1.438 \\
Average tweets per user & 1.2 & 1.4 \\
Average tweet length (words) & 29.20 & 24.37 \\
Shortest tweet length & 5 & 5 \\
Longest tweet length & 72 & 121 \\
%Number of hashtags & 23,170 & 24,444 \\
%Number of URLs & 76,006 & 72,233 \\
%Avg htags per tweet (counting only tweets w/ htags) & 1.817 & 1.406 \\
%Avg url per tweet (counting only tweets w/ urls) & 1.035 & 1.011 \\
Number of tweets with hashtag & 12,751 & 17,390 \\
Number of tweets with URL & 73,439 & 71,434 \\
Number of tweets with emoji & 9,971 & 8,509 \\
\hline
\end{tabular}
\caption{Dataset statistics. }%We construct two large-scale datasets including English (EN) and Turkish (TR) tweets for hate speech detection in terms of three classes (hate, offensive, and normal).}
\label{tab:dataset_stats}
\end{table}

\begin{table}[t]
\small
\centering
\setlength{\tabcolsep}{3pt}
\begin{tabular}{llrrrr}
\hline
\textbf{Lang.} & \textbf{Domain} & \textbf{Hate} & \textbf{Offens.} & \textbf{Normal} & \textbf{Total} \\
\hline
\multirow{5}{*}{EN} & Religion & 1,427 & 5,221 & 13,352 & 20k \\
& Gender & 1,313 & 6,431 & 12,256 & 20k \\
& Race & 1,541 & 3,846 & 14,613 & 20k \\
& Politics & 1,610 & 6,018 & 12,372 & 20k \\
& Sports & 1,434 & 5,624 & 12,942 & 20k \\
\hline
\multirow{5}{*}{TR} & Religion & 5,688 & 7,435 & 6,877 & 20k \\
& Gender & 2,780 & 6,521 & 10,699 & 20k \\
& Race & 5,095 & 4,905 & 10,000 & 20k \\
& Politics & 7,657 & 4,253 & 8,090 & 20k \\
& Sports & 6,373 & 7,633 & 5,994 & 20k \\
\hline
\end{tabular}
\caption{Distribution of tweets in our datasets.}
\label{tab:tweet_distributions}
\end{table}

\section{Experiments}
\label{section:experiments}
We have two main experiments. First, we analyze the performance of various models for hate speech detection. In the second part, we examine cross-domain transfer for hate speech detection.

\subsection{Hate Speech Detection}

\subsubsection{Experimental Design}\label{section:experiments_det_des}

We apply 10-fold cross-validation, where each fold has 90k train instances; and report the average score of weighted precision, recall, and F1 score. Since the dataset is unbalanced, we measure weighted metrics and avoid to report accuracy. The evaluation scores for each class are also examined in the scalability experiments in Section \ref{section:scalability}. We determine statistically significant differences between the means, which follow non-normal distributions, by using the two-sided Mann-Whitney U (MWU) test at \%95 interval with Bonferroni correction. We compare the performances of three family of models. 

\begin{itemize}[leftmargin=*,noitemsep]

\item \textbf{BOW}: We encode tweets using the bag-of-words model (BOW) with TF-IDF term weightings, and train a multinomial Logistic Regression classifier for 1000 iterations, using default parameters with sk-learn \cite{Pedregosa:2011}. TF-IDF term weightings are extracted from the train and test sets separately.

\item \textbf{Neural}: We employ two neural models, CNN \cite{Kim:2014} and LSTM \cite{Hochreiter:1997}, using a dense classification layer on top with cross-entropy loss. For both models, we use Adam optimizer \cite{Kingma:2015} with a learning rate of 5e-5 for 10 epochs. FastText's English and Turkish word embeddings \cite{Grave:2018} are given as input with a dimension size of 300. For CNN, we use 100 kernels each having sizes between 3 and 5. We use PyTorch \cite{Paszke:2019} implementations for both.

\item \textbf{Transformer LM}: We fine-tune Transformer-based language models that are pre-trained on English, Turkish, and multilingual text corpus. We use Huggingface \cite{Wolf:2019} implementation for Transformer-based language models.

\end{itemize}

We fine-tune the following models that are pre-trained by using English or Turkish text:

\begin{itemize}[leftmargin=*,noitemsep]

    \item \textbf{BERT} \cite{Devlin:2019}: BERT uses bi-directional masked language modeling and next sentence prediction.
    
    \item \textbf{BERTweet} \cite{Nguyen:2020}: BERTweet is trained based on the RoBERTa \cite{Liu:2019b} pre-training procedure by using only tweets.
    
    \item \textbf{ConvBERT} \cite{Jiang:2020}: ConvBERT architecture replaces the quadratic time complexity of the self-attention mechanism of BERT with convolutional layers. The number of self-attention heads are reduced by a mixed attention mechanism of self-attention and convolutions that would model local dependencies. 
    
    \item \textbf{Megatron} \cite{Shoeybi:2019}: Megatron introduces an efficient parallel training approach for BERT-like models to increase parameter size.
    
    \item \textbf{RoBERTa} \cite{Liu:2019b}: RoBERTa is built on the BERT architecture with modified hyperparameters and a diverse corpora in pre-training, and removes the task of next sentence prediction.

    \item \textbf{BERTurk} \cite{Schweter:2020}: The model re-trains BERT architecture for Turkish data.
    
    \item \textbf{ConvBERTurk} \cite{Schweter:2020}: Based on ConvBERT \cite{Jiang:2020}, but using a modified training procedure and Turkish data.

\end{itemize}

To understand the generalization capability of multilingual models to both English and Turkish, we fine-tune the following multilingual models.

\begin{itemize}[leftmargin=*,noitemsep]
    \item \textbf{mBERT} \cite{Devlin:2019}: mBERT is built on the BERT architecture, but using multilingual text covering 100 languages.
    
    \item \textbf{XLM-R} \cite{Conneau:2020}: XLM-R is built on the RoBERTa architecture, but using multilingual text covering 100 languages. The model is trained on more data compared to mBERT, and removes the task of next sentence prediction.
\end{itemize}

We apply the same experimental settings to all models. Batch size is 32, learning rate is 1e-5, the number of epochs is 5, maximum input length is 128 tokens, using AdamW optimizer. Only exception is Megatron, due to its large size, we reduce batch size to 8. We use GeForce RTX 2080 Ti for fine-tuning.

\begin{table}[t]
\small
\centering
\setlength{\tabcolsep}{1.5pt}
\begin{tabular}{lccc|ccc}
\hline
\multirow{2}{*}{\textbf{Model}} & \multicolumn{3}{c|}{\textbf{EN}} & \multicolumn{3}{c}{\textbf{TR}} \\
& \textbf{Prec.} & \textbf{Recall} & \textbf{F1} & \textbf{Prec.} & \textbf{Recall} & \textbf{F1} \\
\hline
BOW & 0.777 & 0.796 & 0.779 & 0.707 & 0.710 & 0.706 \\
\hline
CNN & 0.779 & 0.796 & 0.782 & 0.676 & 0.679 & 0.675 \\
LSTM & 0.787 & 0.798 & 0.790 & 0.689 & 0.688 & 0.686 \\
\hline
BERT & 0.815 & 0.817 & 0.816 & - & - & - \\
BERTweet & 0.825 & 0.829 & 0.826 $\circ$ & - & - & - \\
ConvBERT & 0.823 & 0.825 & 0.823 & - & - & - \\
Megatron & \textbf{0.831} & \textbf{0.830} & \textbf{0.830} $\bullet$ & - & - & - \\
RoBERTa & 0.822 & 0.826 & 0.823 & - & - & - \\
\hline
mBERT & 0.817 & 0.818 & 0.818 & 0.757 & 0.752 & 0.753 \\
XLM-R & 0.823 & 0.826 & 0.824 & 0.770 & 0.767 & 0.768 \\
\hline
BERTurk & - & - & - & 0.778 & 0.777 & 0.777 $\circ$ \\
ConvBERTurk & - & - & - & \textbf{0.781} & \textbf{0.782} & \textbf{0.782} $\bullet$ \\
\hline
\end{tabular}
\caption[caption]{\textbf{Multi-class hate speech detection}. Average of 10-fold cross-validation is reported. Highest score is given in bold. Models are divided into sub-groups in terms of BOW, Neural, and Transformer models (English, multilingual, and Turkish language models). The symbol ``$\bullet$" indicates statistical significant difference at a 95\% interval (with Bonferroni correction $p<.006$ for English and $p<.008$ for Turkish) in pairwise comparisons between the highest performing method and others (except the ones with ``$\circ$").}
\label{tab:cross_domain_results}
\end{table}

\subsubsection{Experimental Results}

In Table \ref{tab:cross_domain_results}, we report the performance of multi-class (hate, offensive, and normal) hate speech detection.

\textbf{Transformer-based language models outperform conventional ones in large-scale hate speech detection.} The highest performing models are Megatron with the highest number of model parameters in English, and ConvBERTurk in Turkish. BERTweet has higher performance than BERT, showing the importance of pre-training corpus. Conventional models (BOW, CNN, and LSTM) are not as successful as Transformer-based models in both languages. There are approximately 5\% performance gap between the highest scores by conventional models and Transformer models in English, and 10\% for Turkish.

\textbf{Conventional BOW can be competitive.} We observe that the bag-of-words model has surprisingly, competitive performance in both languages. We note that the tweets in all classes are obtained with the same keyword set. However, a possible reason could be the existence of expressions and keywords that are specific to offensive or hate classes, such as slurs and curse words. 

\textbf{Multilingual language models can be effective for trade-off between performance and language flexibility.} 
Multilingual models, mBERT and XLM-R, have challenging performance with the models pre-trained using only English text. XLM-R has close results to BERTurk (pre-trained with Turkish text) as well. Multilingual models can thereby provide language flexibility (i.e., removing training dependency on new language) without sacrificing substantial task performance. 

\subsection{Cross-Domain Transfer}

\begin{table*}[h]
\small
\centering
\renewcommand{\arraystretch}{1.6}
\def\mathbi#1{\textbf{\em #1}}
\newcommand{\transp}{^{\mathsf{T}}}
\begin{tabular}{llp{0.7cm}p{0.7cm}p{0.7cm}p{0.7cm}p{0.7cm}||p{0.7cm}p{0.7cm}p{0.7cm}p{0.7cm}p{0.7cm}|
p{0.7cm}}
\hline
& \textbf{Source/Target} & RE & GE & RA & PO & SP & RE & GE & RA & PO & SP & Avg. \\
\hline 
\multirow{5}{*}{EN} & Religion & \cellcolor[HTML]{C0C0C0} 0.804 & \cellcolor[HTML]{6495ED} 92\% & \cellcolor[HTML]{3F61FF} 96\% & \cellcolor[HTML]{3F61FF} 98\% & \cellcolor[HTML]{3F61FF} 97\% & \cellcolor[HTML]{C0C0C0} 0.804 & \cellcolor[HTML]{ffb2b2} -9\% & \cellcolor[HTML]{ffd1d1} -1\% & \cellcolor[HTML]{ffd1d1} -2\% & \cellcolor[HTML]{ffffff} 0\% &  -3\% 
 \\
& Gender & \cellcolor[HTML]{1F51FF} 101\% & \cellcolor[HTML]{C0C0C0} 0.799 & \cellcolor[HTML]{3F61FF} 98\% & \cellcolor[HTML]{3F61FF} 99\% & \cellcolor[HTML]{3F61FF} 99\% & \cellcolor[HTML]{ffffff} 0\% & \cellcolor[HTML]{C0C0C0} 0.799 & \cellcolor[HTML]{ffffff} 0\% & \cellcolor[HTML]{ffffff} 0\% & \cellcolor[HTML]{ffffff} 0\% &  0\% 
\\ 
& Racism & \cellcolor[HTML]{3F61FF} 99\% & \cellcolor[HTML]{6495ED} 93\% & \cellcolor[HTML]{C0C0C0} 0.823 & \cellcolor[HTML]{3F61FF} 96\% & \cellcolor[HTML]{6495ED} 94\% & \cellcolor[HTML]{ffd1d1} -3\% & \cellcolor[HTML]{ff7b7b} -10\% & \cellcolor[HTML]{C0C0C0} 0.823 & \cellcolor[HTML]{ffb2b2} -6\% & \cellcolor[HTML]{ffd1d1} -2\% &  -5\% \\ 
& Politics & \cellcolor[HTML]{3F61FF} 97\% & \cellcolor[HTML]{87CEFF} 89\% & \cellcolor[HTML]{3F61FF} 95\% & \cellcolor[HTML]{C0C0C0} 0.808 & \cellcolor[HTML]{3F61FF} 98\% & \cellcolor[HTML]{ffd1d1} -3\% & \cellcolor[HTML]{ff7b7b} -12\% & \cellcolor[HTML]{ffd1d1} -3\% & \cellcolor[HTML]{C0C0C0} 0.808 & \cellcolor[HTML]{ffffff} 0\% &  -5\% \\ 
& Sports & \cellcolor[HTML]{3F61FF} 95\% & \cellcolor[HTML]{6495ED} 90\% & \cellcolor[HTML]{6495ED} 93\% & \cellcolor[HTML]{3F61FF} 99\% & \cellcolor[HTML]{C0C0C0} 0.853 & \cellcolor[HTML]{ff7b7b} -10\% & \cellcolor[HTML]{ff5050} -15\% & \cellcolor[HTML]{ff7b7b} -11\% &\cellcolor[HTML]{ffb2b2} -6\% & \cellcolor[HTML]{C0C0C0} 0.853 & -11\% \\ 
\cline{2-7}
% & Avg. & \cellcolor[HTML]{3F61FF} 98\% & \cellcolor[HTML]{6495ED} 91\% & \cellcolor[HTML]{3F61FF} 96\% & \cellcolor[HTML]{3F61FF} 98\% & \cellcolor[HTML]{3F61FF} 97\% \\
& Avg. &  98\% & 91\% & 96\% &  98\% & 97\% \\
\hline
\hline
\multirow{5}{*}{TR} & Religion & \cellcolor[HTML]{C0C0C0} 0.754 & \cellcolor[HTML]{6495ED} 93\% & \cellcolor[HTML]{3F61FF} 96\% & \cellcolor[HTML]{6495ED} 93\% & \cellcolor[HTML]{3F61FF} 95\% & \cellcolor[HTML]{C0C0C0} 0.754 & \cellcolor[HTML]{ffb2b2} -5\% & \cellcolor[HTML]{ffd1d1} -1\% & \cellcolor[HTML]{ffb2b2} -6\% & \cellcolor[HTML]{ffffff} 0\% &  -3\% \\ 
& Gender & \cellcolor[HTML]{6495ED} 94\% & \cellcolor[HTML]{C0C0C0} 0.772 & \cellcolor[HTML]{3F61FF} 95\% & \cellcolor[HTML]{87CEFF} 89\% & \cellcolor[HTML]{6495ED} 94\% & \cellcolor[HTML]{ffb2b2} -8\% & \cellcolor[HTML]{C0C0C0} 0.772 & \cellcolor[HTML]{ffd1d1} -4\% & \cellcolor[HTML]{ff7b7b} -12\% & \cellcolor[HTML]{ffd1d1} -3\% & -7\% \\ 
& Racism & \cellcolor[HTML]{3F61FF} 97\% & \cellcolor[HTML]{6495ED} 94\% & \cellcolor[HTML]{C0C0C0}0.779 & \cellcolor[HTML]{6495ED} 92\% & \cellcolor[HTML]{3F61FF} 95\% & \cellcolor[HTML]{ffb2b2} -6\% & \cellcolor[HTML]{ffb2b2} -7\% & \cellcolor[HTML]{C0C0C0} 0.779 & \cellcolor[HTML]{ffb2b2} -9\% & \cellcolor[HTML]{ffd1d1} -2\% &  -6\% \\ 
& Politics & \cellcolor[HTML]{6495ED} 90\% & \cellcolor[HTML]{87CEFF} 89\% & \cellcolor[HTML]{6495ED} 92\% & \cellcolor[HTML]{C0C0C0} 0.765 & \cellcolor[HTML]{6495ED} 90\% & \cellcolor[HTML]{ff7b7b} -11\% & \cellcolor[HTML]{ff7b7b} -10\% & \cellcolor[HTML]{ffb2b2} -6\% &\cellcolor[HTML]{C0C0C0} 0.765 & \cellcolor[HTML]{ffb2b2} -6\% &  -8\% \\ 
& Sports & \cellcolor[HTML]{6495ED} 92\% & \cellcolor[HTML]{87CEFF} 86\% & \cellcolor[HTML]{6495ED} 91\% & \cellcolor[HTML]{B7EFFB} 84\% & \cellcolor[HTML]{C0C0C0} 0.799 & \cellcolor[HTML]{ff7b7b} -13\% & \cellcolor[HTML]{ff5050} -17\% & \cellcolor[HTML]{ff7b7b} -11\% & \cellcolor[HTML]{ff5050} -19\% & \cellcolor[HTML]{C0C0C0} 0.799 &  -15\% \\ 
\cline{2-7}
% & Avg. & \cellcolor[HTML]{6495ED} 93\% & \cellcolor[HTML]{6495ED} 91\% & \cellcolor[HTML]{6495ED} 94\% & \cellcolor[HTML]{6495ED} 90\% & \cellcolor[HTML]{6495ED} 94\% \\
& Avg. &  93\% &  91\% &  94\% & 90\% & 94\% \\
\hline 
\end{tabular}
\caption{Cross-domain transfer for hate speech detection in terms of \textbf{column-wise recovery ratio} and \textbf{row-wise decay ratio}. Source domains are given in rows, targets in columns. The diagonal gray cells have the weighted F1 scores when source and target domains are the same. Recovery scores should be interpreted column-wise, e.g. 92\% recovery from religion to gender in EN means that we recover 92\% of 0.799 (gender to gender). %, but not 0.804 (religion to religion). 
As recovery increases, blue color gets darker. Decay scores should be interpreted row-wise, e.g. -9\% decay from religion to gender in EN means that we lose 9\% of 0.804 (religion to religion). %, but not 0.799 (gender to gender). 
If there is no loss in performance, decay is zero. As decay increases, red color gets darker.}
\label{tab:decay_recovery_merged}
\end{table*}

\subsubsection{Experimental Design}
We examine cross-domain transfer by fine-tuning the model on a source domain, and evaluating on a target domain. The performance can be measured by relative zero-shot transfer ability \cite{Turc:2021}. We refer to it as \emph{recovery ratio}, since it represents the ratio of how much performance is recovered by changing source domain, given as follows.

\begin{equation}
recovery(S,T) = \frac{M(S,T)}{M(T,T)}
\label{eq:recovery}
\end{equation}

\noindent where $M(S,T)$ is a model performance for the source domain $S$ on the target domain $T$. For the recovery ratio, we set a hate domain as target, and remaining ones as source. When source and target domains are the same, recovery would be 1.0.

We also adapt the measurement used in cross-lingual transfer gap \cite{Hu:2020}.
% The performance can also be measured by cross-lingual transfer gap \cite{Hu:2020}. 
We modify it to normalize, and refer to it as \emph{decay ratio}, since it represents the ratio of how much performance is decayed by replacing target domain, given as follows.

\begin{equation}
decay(S,T) = \frac{M(S,T)-M(S,S)}{M(S,S)}
 \label{eq:decay}
\end{equation}

For the decay ratio, we set a hate domain as source, and remaining ones as target. In the case that source and target domains are the same, there would be no decay or performance drop, so decay would be zero. In the cross-domain experiments, we measure weighted F1; and use BERT for English, and BERTurk for Turkish with the same hyperparameters used in Section \ref{section:experiments_det_des}. We apply 10-fold cross-validation, where each fold has 18k train instances in a particular hate domain; and report the average score of recovery and decay in 2k test instances of the corresponding hate domain.

\subsubsection{Experimental Results}
The recovery and decay scores are given in Table \ref{tab:decay_recovery_merged}. %, and the decay scores in Table \ref{tab:decay}. 
We note that recovery and decay represent independent measures for domain transfer performance.  For instance, in English, the domain transfer from gender to politics has 99\% recovery, and its decay ratio is 0\%. The domain transfer from sports to politics has the same recovery ratio, but its decay is -6\%, which shows that the same recovery values do not necessarily mean the same performance.

\textbf{Hate domains can mostly recover each other's performance.} Recovery performances between domains are quite effective, such that, on average, 96\% of the performance of a target domain is recovered by others for English, and 92\% for Turkish. The training dataset composed of only a single domain can be thereby employed to detect hate speech patterns of another domain.  %When the number of datasets for hate speech detection in the literature is considered, it may be possible to use existing datasets with different domains as source without curating new training datasets.
We argue that there can be overlapping hate patterns across multiple domains, which can be examined for hate speech in a more fundamental and social level. Moreover, common vocabulary across different topics can introduce domain transitivity such as women's sports or women in politics.

\textbf{Recovering gender is more difficult than other domains.} Gender-based hate tweets can not be easily predicted by other hate domains, as the average recover ratio for gender is lower than others, 91\% in both English and Turkish. In addition to gender, politics has the average recover ratio of 90\% in Turkish. One can deduce that hate speech patterns of these domains display different characteristics from general hate patterns. 

\textbf{Sports cannot generalize to other domains.} While sports can be recovered by other domains, the average decay ratio of sports is poor (more than 10\%) in both languages, as observed in Table \ref{tab:decay_recovery_merged}. A possible reason can be specific slang language used in sports.

\textbf{Gender can generalize to other domains in English, but not in Turkish.} Gender can maintain its in-domain success to other domains in English, as its average decay ratio is zero. Although average decay ratio is not too high in Turkish, it is still higher than English. One possible reason could be that Turkish has a gender neutral grammar with gender-free pronouns. The success of gender in English can be important for data scarcity in hate speech detection, since one can use the model trained with gender instances to infer a target domain.

\section{Discussion}\label{section:discussion}

\subsection{Error Analysis}
We provide an error analysis on the model predictions in Table \ref{tab:error_analysis}. We select a representative model from each model family that we compare in the previous section; namely BERT/BERTurk, XLM-R, CNN, and BOW. 

There are eight tweet examples divided into three groups. The first group with the tweets numbered 1 to 4 is given to represent the examples of both success and failure of Transformer-based language models. BERT and XLM-R incorrectly predict the first tweet, possibly due to giving more attention to ``I hate my life" that is not actually a hateful phrase but describes dislike for the current situation. CNN's prediction is also incorrect possibly due to convolution on the same phrase. BOW's prediction is correct by using a sparse vector with many non-hate words. The second tweet is a similar example given for Turkish that BERT and XLM-R probably give more attention to ``almamız gereken bir intikam var" (translated as ``there is a revenge we need to take"). On the other hand, BERT and XLM-R succeed in the third and fourth tweets due to ``I'll kill them" and ``bir kaç erkek ölsün istiyorum" (translated as ``I want a few men to die"). The difference from the first two examples is that the third and fourth tweets include true-hate phrases.  

The second group is given to show the model performance on offensive tweets. All models except XLM-R incorrectly label the fifth tweet as offensive, possibly due to the existence of ``f**g". On the other hand, BERT and XLM-R correctly predict the sixth tweet. There are no clear offensive words in this tweet. CNN and BOW therefore fail in this tweet, while BERT and XLM-R could capture the semantics. 

\begin{table*}[t]
\small
\centering
\renewcommand{\arraystretch}{0.9}
\begin{tabular}{lllcccc}
\hline
\multirow{2}{*}{\textbf{\#}} & \multirow{2}{*}{\textbf{Label}} & \multirow{2}{*}{\textbf{Tweet}} & \multicolumn{4}{c}{\textbf{Predictions}}\\
& & & \textbf{BERT} & \textbf{XLM-R} & \textbf{CNN} & \textbf{BOW} \\
\hline
\multirow{2}{*}{1} & \multirow{2}{*}{Normal} & Was on an absolute heater, then I lose 500 dollars on & \multirow{2}{*}{Hate} & \multirow{2}{*}{Hate} & \multirow{2}{*}{Hate} & \multirow{2}{*}{\textbf{Normal}} \\
& & Korean baseball and Costa Rican soccer. I hate my life & & & \vspace{0.1cm} \\
\multirow{3}{*}{2} & \multirow{3}{*}{Normal} & Kazanmamız gereken bir maç, almamız gereken bir  & \multirow{3}{*}{Hate} & \multirow{3}{*}{Hate} & \multirow{3}{*}{\textbf{Normal}} & \multirow{3}{*}{\textbf{Normal}} \\
& & intikam var. Allah büyük. (We have a game to win,& & \\
& &  a revenge to take. God is great.) & & \vspace{0.1cm} \\
\multirow{2}{*}{3} & \multirow{2}{*}{Hate} & ...I'm that fast. Hitting someone going full speed. & \multirow{2}{*}{\textbf{Hate}} & \multirow{2}{*}{\textbf{Hate}} & \multirow{2}{*}{Normal} & \multirow{2}{*}{Normal} \\ 
& & Over 20 mph 190 lbs. I'll kill them & & &  &  \vspace{0.1cm} \\ 
\multirow{3}{*}{4} & \multirow{3}{*}{Hate} & Feminist değilim de, tanıdığıma pişman olduğum bir kaç & \multirow{3}{*}{\textbf{Hate}} & \multirow{3}{*}{\textbf{Hate}} & \multirow{3}{*}{Normal}  & \multirow{3}{*}{Normal} \\
& & erkek ölsün istiyorum (I am not a feminist, but I want & & & \\ 
& & a few men to die that I regret knowing)& & & \\ 
\hline
\multirow{2}{*}{5} & \multirow{2}{*}{Normal} & I’m gay but I’ve got the biggest fucking crush on & \multirow{2}{*}{Offensive} & \multirow{2}{*}{\textbf{Normal}} & \multirow{2}{*}{Offensive} & \multirow{2}{*}{Offensive} \\
& & @katyperry I have for years! &  & & & \vspace{0.1cm}  \\ 
\multirow{2}{*}{6} & \multirow{2}{*}{Offensive} & The three different ways to become a brain dead zombie & \multirow{2}{*}{\textbf{Offensive}} & \multirow{2}{*}{\textbf{Offensive}} & \multirow{2}{*}{Normal} & \multirow{2}{*}{Normal} \\
& & are, a virus, radiation and Christianity. & & & & \\ 
\hline
7 & Hate & A gay killing another gay nice
& Normal & Normal & Normal & Normal \vspace{0.1cm}  \\
\multirow{2}{*}{8} & \multirow{2}{*}{Hate} & Yahudiler cinayet Araplar hiyanet kavmidir & \multirow{2}{*}{Normal} & \multirow{2}{*}{Normal} & \multirow{2}{*}{Normal} & \multirow{2}{*}{Normal} \\
& & (Jews are a people of murder, Arabs of treason) &  &  &  \\ 
\hline
\end{tabular}
\caption{Error analysis using model predictions. Correct predictions are given in bold. For Turkish tweet examples, BERT refers to BERTurk. Translations for Turkish tweets are given in parentheses.}
\label{tab:error_analysis}
\end{table*}

The last group is given to show hard examples that all models fail. The seventh tweet is difficult to detect, since the positive word ``nice" can be confusing in this short tweet. %The seventh tweet is a very short example written in social media language, which makes hateful semantics hard to detect (we further check that the prediction of BERTweet is offensive for this tweet). 
The models fail to understand the semantics in the last Turkish tweet.  

\subsection{Existing Models}

We evaluate the generalization capability of existing hate speech detection models to our dataset. Since there is no publicly available Turkish-specific models, we only examine it in English. We use the following fine-tuned models for zero-shot inference, as well as fine-tuning them further with our data. 

\begin{itemize}[leftmargin=*,noitemsep]
    \item \textbf{HateXplain} \cite{Mathew:2021}: HateXplain fine-tunes BERT, using a novel dataset with 20k instances, 9k of which are tweets. The model can be used for zero-shot inference on multi-class (hate, offensive, and normal) detection.
    \item \textbf{HateBERT} \cite{Caselli:2021}: HateBERT pre-trains BERT architecture by using approximately 1.5m Reddit messages by suspended communities due to promoting hateful content. The model can be used for zero-shot inference on binary classification (hate or not). To adapt our dataset to binary classification, we merge offensive and hate tweets.
\end{itemize}

We apply 10-fold cross-validation, as in previous experiments. The results of existing models are given Table \ref{tab:binary_zero}. We compare their performances on our dataset with BERT's performance. In multi-class scheme, HateXplain cannot outperform BERT in both zero-shot inference and further fine-tune scheme. In binary scheme, zero-shot inference fails more compared to HateXplain despite having less number of classes. The poor performance of zero-shot HateBERT is probably due to Reddit messages used in pre-training, while our dataset consists of tweets. %Fine-tuning the model further provides significant increase in performance, having similar performance with BERT.
Fine-tuning the model further provides similar performance with BERT. Overall, we show that existing models have limited generalization capability to new data. A possible reason for existing models failing to generalize can be that our dataset consists of tweets from various topics.

\begin{table}[t]
\small
\centering
\begin{tabular}{lcc|cc}
\hline
\multirow{2}{*}{\textbf{Model}} & \multicolumn{2}{c|}{\textbf{Mutliclass}} & \multicolumn{2}{c}{\textbf{Binary}} \\
& \textbf{F1} & \textbf{Type} & \textbf{F1} & \textbf{Type} \\
\hline
BERT & 0.816 & Fine-tune & 0.862 & Fine-tune \\
HateXplain & 0.796 & Fine-tune & - & - \\
HateXplain & 0.769 & Zero-shot & - & - \\
HateBERT & - & - & 0.865 & Fine-tune \\
HateBERT & - & - & 0.485 & Zero-shot  \\
\hline
\end{tabular}
\caption{Zero-shot and further fine-tuning results for existing hate speech detection models.}
\label{tab:binary_zero}
\end{table}

\begin{figure*}[ht]
    \centering
    \begin{subfigure}[t]{0.32\textwidth}
        \centering
        \includegraphics[width=\textwidth]{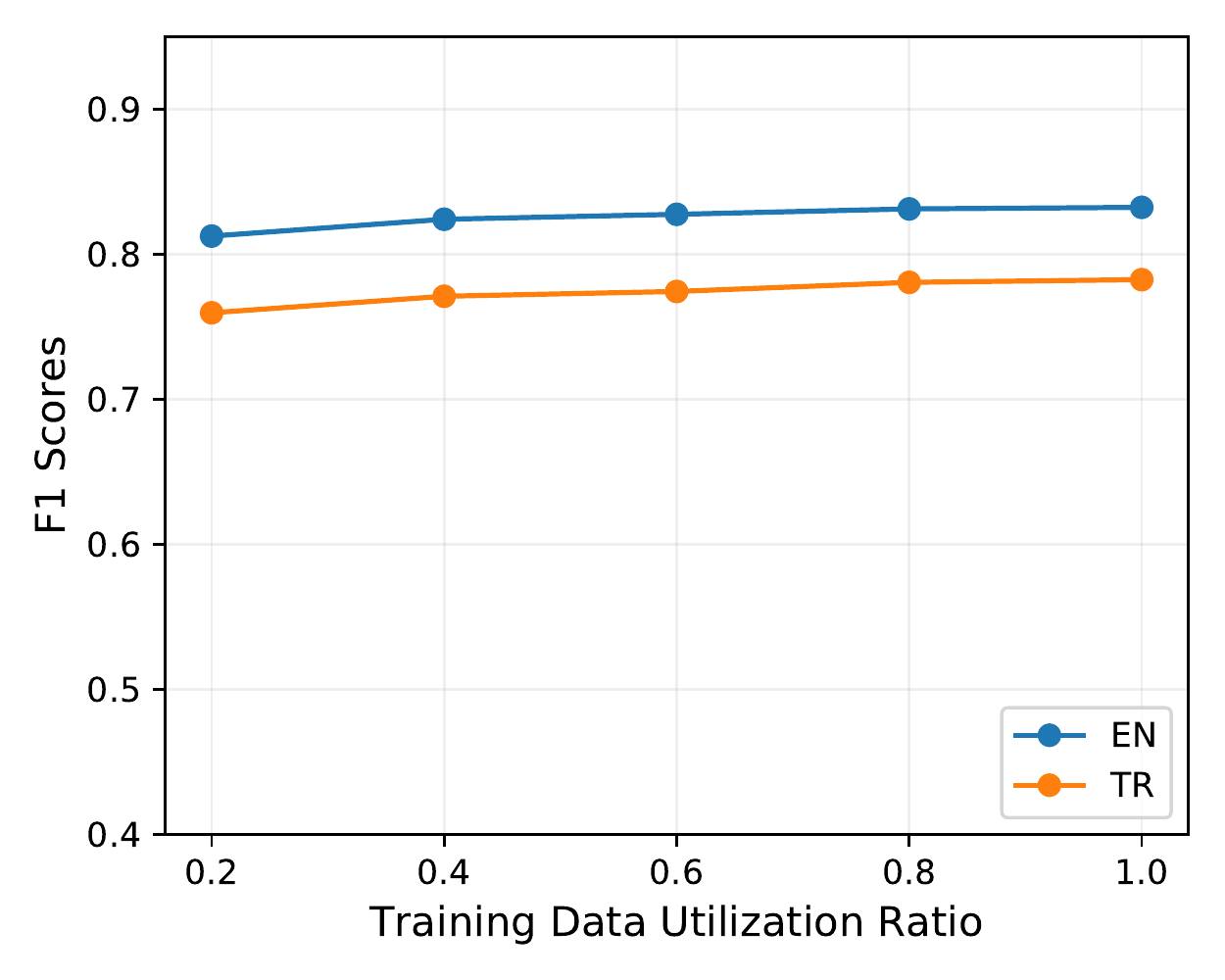}
        \caption{Weighted F1 scores for multi-class hate speech detection for different scales of training data. There is a slight performance increase in both languages.}
        \label{fig:scale_load_best}
     \end{subfigure}
     \hfill
     \begin{subfigure}[t]{0.32\textwidth}
        \centering
        \includegraphics[width=\textwidth]{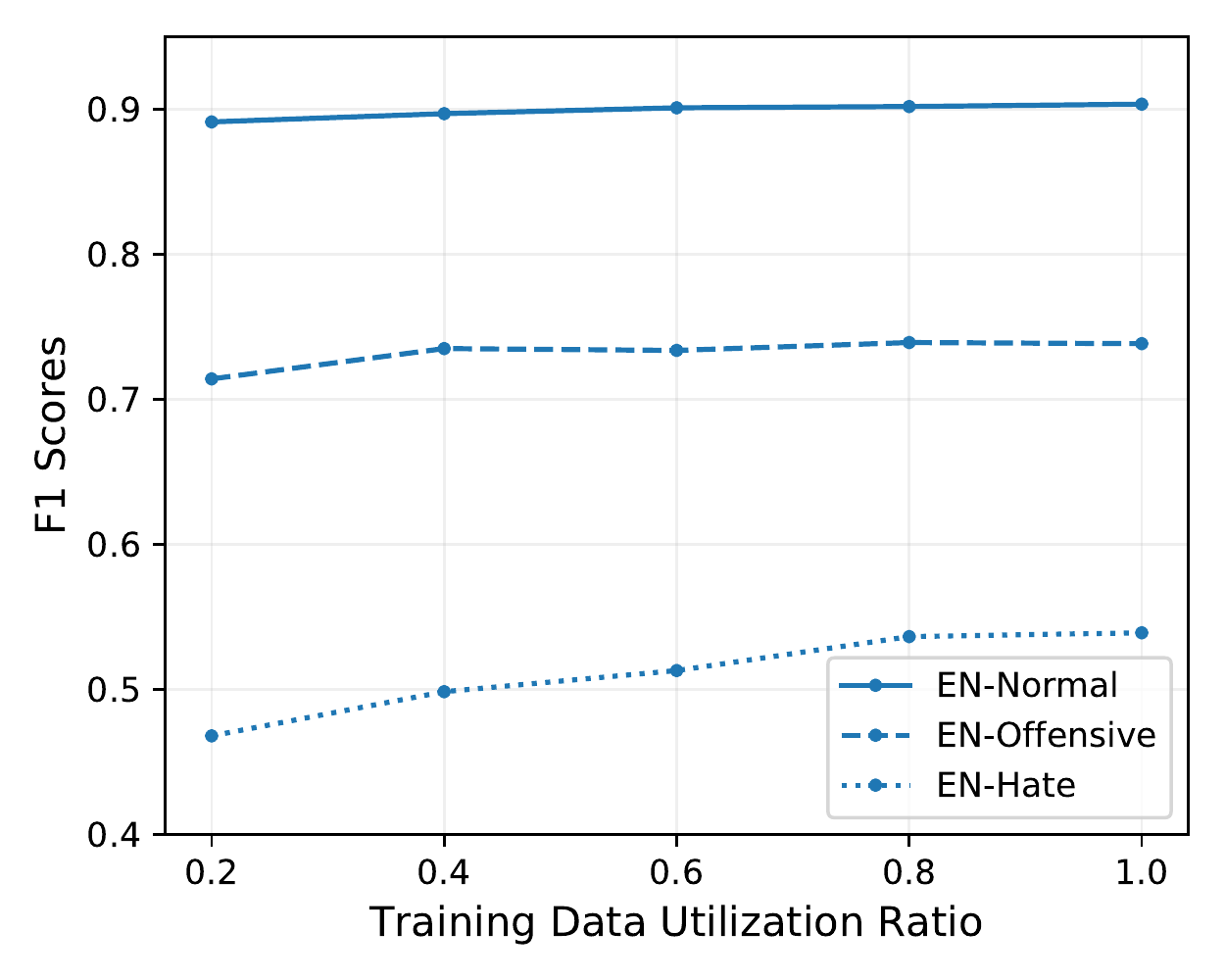}
        \caption{Weighted F1 scores for different classes in EN. The performance of normal class saturates early, and hate class benefits the most.}
        \label{fig:en_class_load_best}
     \end{subfigure}
     \hfill
     \begin{subfigure}[t]{0.32\textwidth}
        \centering
        \includegraphics[width=\textwidth]{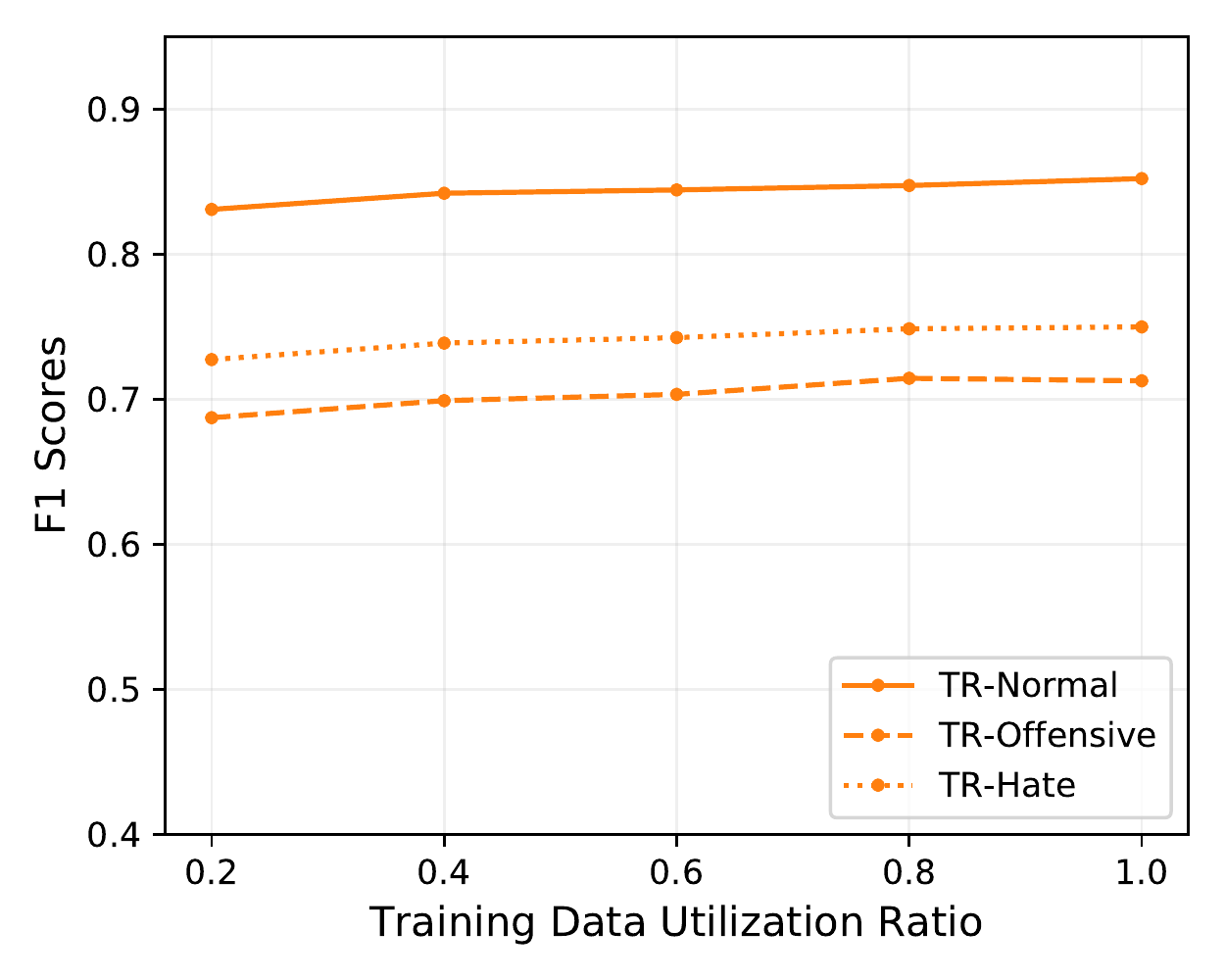}
        \caption{Weighted F1 scores for different classes in TR. There is a slight performance increase in all classes.}
        \label{fig:tr_class_load_best}
     \end{subfigure}
        \caption{Scalability analysis for hate speech detection.}
        \label{fig:scalability_plots}
\end{figure*}

\subsection{Scalability}
\label{section:scalability}
We examine scalability as the effect of increasing training size on model performance. Since labeling hate speech data is costly, the data size of hate speech detection becomes important. Our large-scale datasets are available to analyze scalability. To do so, we split 10\% of data for testing, 10\% for validation, and remaining 80\% for training. From the training split, we set five scale values starting from 20\% to 100\%. To obtain reliable results, we repeat this process five times, and report the average scores. At each iteration, training and validation datasets are randomly sampled. We use BERT for English, and BERTurk for Turkish with the same hyperparameters used in Section \ref{section:experiments_det_des}.

We train the models for five epochs, but report the highest performance during training to have a fair comparison by neglecting the positive effect of having more training data, since more number of instances means more number of train steps. We observe that using smaller number of instances (e.g. 20\% of data size) needs more epochs to converge, compared to larger data. The highest performances are obtained for 20-60\% scales at 2.5 epochs in English and 3.5 epochs in Turkish, whereas for 80-100\% at 2 epochs in English and 2.5 epochs in Turkish.

The results for overall detection performance are given in Figure \ref{fig:scale_load_best}. We observe that the performance slightly improves as training data increases in both English and Turkish. Moreover, 98\% of full-sized data performance can be obtained when 20\% of training instances are used in English. This ratio is 97\% in Turkish. In order to reveal the reason of this result, we also investigate the scalability performance of individual classes in Figure \ref{fig:en_class_load_best} for English, and Figure \ref{fig:tr_class_load_best} for Turkish. 

The model for English has the highest performance in the normal class, and worst in the hate class. Interestingly, the performance of hate class improves significantly as training data increases whereas normal and offensive tweets exhibit a slightly increasing pattern. One can observe the impact of class imbalance on performance improvements. The normal class with more number of instances is not affected much by different scales. The performance improvement is more prominent in the hate class, which has a smaller number of data instances than the normal class. This result emphasizes the importance of the data size, especially number of hate instances, for hate speech detection. Given that the main bottleneck in hate speech detection is misprediction of hate tweets rather than normal ones, using a higher number of data instances with hate content can improve the performance.

The performance of all classes slightly increase in Turkish. The performance of predicting hate tweets is higher than offensive ones (vice versa in English). The reason could be the different speech patterns in different languages. Moreover, the hate performance of English is still worse than Turkish when similar number of training instances are considered, e.g., hate score of ratio 100\% in Figure \ref{fig:en_class_load_best} (7,325 hate tweets) is still worse than the score of 20\% in Figure \ref{fig:tr_class_load_best} (5,519 hate tweets). When greater class imbalance in English is considered in this case, we argue that class imbalance is also an important factor besides the number of hate tweets.

\begin{table}[t]
\small
\centering
\begin{tabular}{llccc}
\hline
\textbf{Data} & \textbf{Model} & \textbf{Prec.} & \textbf{Recall} & \textbf{F1}\\
\hline
\multirow{5}{*}{EN} & Raw text & 0.815 & 0.817 & 0.816 \\
& w/o URL & 0.816 & 0.819 & 0.817 \\
& w/o Hashtag & 0.816 & 0.818 & 0.817 \\
& w/o Emoji & 0.815 & 0.818 & 0.816 \\
& w/o Any & 0.817 & 0.818 & 0.817 \\
\hline
\multirow{5}{*}{TR} & Raw text & 0.778 & 0.777 & 0.777 \\
& w/o URL & 0.779 & 0.778 & 0.778 \\
& w/o Hashtag & 0.777 & 0.776 & 0.776 \\
& w/o Emoji & 0.779 & 0.778 & 0.778 \\
& w/o Any & 0.777 & 0.777 & 0.777 \\
\hline
\end{tabular}
\caption{Effect of removing tweet-specific components in terms of the average of 10-fold cross-validation.}
\label{table:ablation}
\end{table}

\subsection{Ablation Study}
To assess the effect of tweet-specific components on the performance of hate speech detection, we remove each component from tweets, and re-run BERT for English, and BERTurk for Turkish. Tweet-specific components that we examine in the ablation study are URLs, hashtags, and emoji symbols. Table \ref{table:ablation} reports the experimental results of the ablation study. The results show that removing tweet-specific components has almost no effect on the performances for both languages. The reason could be that the numbers of hashtags and emojis are low in the dataset, as observed in Table \ref{tab:dataset_stats}. On the other hand, there are many tweets with URLs, yet there is no significant difference when URLs are removed. We argue that BERT-like models can be robust to noise in text caused by URLs.

\section{Conclusion} \label{section:conclusion}

We construct large-scale datasets for hate speech detection in English and Turkish to analyze the performances of state-of-the-art models, along with scalability. We design our datasets to have equal size of instances for each of five hate domains and two languages; so that we can report zero-shot cross-domain results. We find that Transformer-based language models outperform conventional models, and their performances can be scalable to different training sizes. We also show that the vast majority of the performance of a target domain can be recovered by other domains. In future work, the generalization capability of domains can be examined in other languages and platforms. One can further analyze model scalability beyond Transformer-based language models.

% \nocite{*}
\section{Bibliographical References}\label{reference}
%\label{main:ref}

\bibliographystyle{lrec2022-bib}
\bibliography{references}

\begin{thebibliography}{}

\bibitem[\protect\citename{Agrawal and Awekar}2018]{Agrawal:2018}
Agrawal, S. and Awekar, A.
\newblock (2018).
\newblock Deep learning for detecting cyberbullying across multiple social
  media platforms.
\newblock In {\em Advances in Information Retrieval}, pages 141--153, Cham.
  Springer International Publishing.

\bibitem[\protect\citename{Albadi \bgroup et al.\egroup }2018]{Albadi:2018}
Albadi, N., Kurdi, M., and Mishra, S.
\newblock (2018).
\newblock Are they our brothers? {A}nalysis and detection of religious hate
  speech in the {A}rabic {T}wittersphere.
\newblock In {\em 2018 IEEE/ACM International Conference on Advances in Social
  Networks Analysis and Mining (ASONAM)}, pages 69--76.

\bibitem[\protect\citename{Arango \bgroup et al.\egroup }2019]{Arango:2019}
Arango, A., P\'{e}rez, J., and Poblete, B.
\newblock (2019).
\newblock Hate speech detection is not as easy as you may think: A closer look
  at model validation.
\newblock In {\em Proceedings of the 42nd International ACM SIGIR Conference on
  Research and Development in Information Retrieval}, SIGIR'19, pages 45--54,
  New York, NY, USA. Association for Computing Machinery.

\bibitem[\protect\citename{Basile \bgroup et al.\egroup }2019]{Basile:2019}
Basile, V., Bosco, C., Fersini, E., Nozza, D., Patti, V., Rangel~Pardo, F.~M.,
  Rosso, P., and Sanguinetti, M.
\newblock (2019).
\newblock {S}em{E}val-2019 {T}ask 5: Multilingual detection of hate speech
  against immigrants and women in {T}witter.
\newblock In {\em Proceedings of the 13th International Workshop on Semantic
  Evaluation}, pages 54--63, Minneapolis, Minnesota, USA. Association for
  Computational Linguistics.

\bibitem[\protect\citename{Byman}2021]{Byman:2021}
Byman, D.~L.
\newblock (2021).
\newblock {How hateful rhetoric connects to real-world violence}.
\newblock
  https://www.brookings.edu/blog/order-from-chaos/2021/04/09/how-hateful-rhetoric-connects-to-real-world-violence/.
\newblock Accessed: 2022-01-11.

\bibitem[\protect\citename{Cao \bgroup et al.\egroup }2020]{Cao:2020}
Cao, R., Lee, R. K.-W., and Hoang, T.-A.
\newblock (2020).
\newblock Deep{H}ate: Hate speech detection via multi-faceted text
  representations.
\newblock In {\em 12th ACM Conference on Web Science}, WebSci '20, pages
  11--20, New York, NY, USA. Association for Computing Machinery.

\bibitem[\protect\citename{Caselli \bgroup et al.\egroup }2021]{Caselli:2021}
Caselli, T., Basile, V., Mitrovi{\'c}, J., and Granitzer, M.
\newblock (2021).
\newblock {H}ate{BERT}: Retraining {BERT} for abusive language detection in
  {E}nglish.
\newblock In {\em Proceedings of the 5th Workshop on Online Abuse and Harms
  (WOAH 2021)}, pages 17--25, Online. Association for Computational
  Linguistics.

\bibitem[\protect\citename{Chatzakou \bgroup et al.\egroup
  }2017]{Chatzakou:2017}
Chatzakou, D., Kourtellis, N., Blackburn, J., De~Cristofaro, E., Stringhini,
  G., and Vakali, A.
\newblock (2017).
\newblock Mean birds: Detecting aggression and bullying on {T}witter.
\newblock In {\em Proceedings of the 2017 ACM on Web Science Conference},
  WebSci '17, pages 13--22, New York, NY, USA. Association for Computing
  Machinery.

\bibitem[\protect\citename{Conneau \bgroup et al.\egroup }2020]{Conneau:2020}
Conneau, A., Khandelwal, K., Goyal, N., Chaudhary, V., Wenzek, G.,
  Guzm{\'{a}}n, F., Grave, E., Ott, M., Zettlemoyer, L., and Stoyanov, V.
\newblock (2020).
\newblock Unsupervised cross-lingual representation learning at scale.
\newblock In {\em Proceedings of the 58th Annual Meeting of the Association for
  Computational Linguistics}, pages 8440--8451. Association for Computational
  Linguistics.

\bibitem[\protect\citename{Davidson \bgroup et al.\egroup }2017]{Davidson:2017}
Davidson, T., Warmsley, D., Macy, M., and Weber, I.
\newblock (2017).
\newblock Automated hate speech detection and the problem of offensive
  language.
\newblock In {\em Proceedings of the International AAAI Conference on Web and
  Social Media}, volume~11, pages 512--515.

\bibitem[\protect\citename{Devlin \bgroup et al.\egroup }2019]{Devlin:2019}
Devlin, J., Chang, M.-W., Lee, K., and Toutanova, K.
\newblock (2019).
\newblock {BERT}: Pre-training of deep bidirectional transformers for language
  understanding.
\newblock In {\em Proceedings of the 2019 Conference of the North American
  Chapter of the Association for Computational Linguistics: Human Language
  Technologies, Volume 1 (Long and Short Papers)}, pages 4171--4186,
  Minneapolis, Minnesota. Association for Computational Linguistics.

\bibitem[\protect\citename{Fernquist \bgroup et al.\egroup
  }2019]{Fernquist:2019}
Fernquist, J., Lindholm, O., Kaati, L., and Akrami, N.
\newblock (2019).
\newblock A study on the feasibility to detect hate speech in {S}wedish.
\newblock In {\em 2019 IEEE International Conference on Big Data}, pages
  4724--4729.

\bibitem[\protect\citename{Fi{\v{s}}er \bgroup et al.\egroup }2017]{Fiser:2017}
Fi{\v{s}}er, D., Erjavec, T., and Ljube{\v{s}}i{\'c}, N.
\newblock (2017).
\newblock Legal framework, dataset and annotation schema for socially
  unacceptable online discourse practices in {S}lovene.
\newblock In {\em Proceedings of the First Workshop on Abusive Language
  Online}, pages 46--51, Vancouver, BC, Canada. Association for Computational
  Linguistics.

\bibitem[\protect\citename{Founta \bgroup et al.\egroup }2018]{Founta:2018}
Founta, A., Djouvas, C., Chatzakou, D., Leontiadis, I., Blackburn, J.,
  Stringhini, G., Vakali, A., Sirivianos, M., and Kourtellis, N.
\newblock (2018).
\newblock Large scale crowdsourcing and characterization of {T}witter abusive
  behavior.
\newblock In {\em International AAAI Conference on Web and Social Media},
  volume~12, pages 491--500.

\bibitem[\protect\citename{Geva \bgroup et al.\egroup }2019]{Geva:2019}
Geva, M., Goldberg, Y., and Berant, J.
\newblock (2019).
\newblock Are we modeling the task or the annotator? {A}n investigation of
  annotator bias in natural language understanding datasets.
\newblock In {\em Proceedings of the 2019 Conference on Empirical Methods in
  Natural Language Processing and the 9th International Joint Conference on
  Natural Language Processing (EMNLP-IJCNLP)}, pages 1161--1166, Hong Kong,
  China. Association for Computational Linguistics.

\bibitem[\protect\citename{Golbeck \bgroup et al.\egroup }2017]{Golbeck:2017}
Golbeck, J., Ashktorab, Z., Banjo, R.~O., Berlinger, A., Bhagwan, S., Buntain,
  C., Cheakalos, P., Geller, A.~A., Gergory, Q., Gnanasekaran, R.~K.,
  Gunasekaran, R.~R., Hoffman, K.~M., Hottle, J., Jienjitlert, V., Khare, S.,
  Lau, R., Martindale, M.~J., Naik, S., Nixon, H.~L., Ramachandran, P., Rogers,
  K.~M., Rogers, L., Sarin, M.~S., Shahane, G., Thanki, J., Vengataraman, P.,
  Wan, Z., and Wu, D.~M.
\newblock (2017).
\newblock A large labeled corpus for online harassment research.
\newblock In {\em Proceedings of the 2017 ACM on Web Science Conference},
  WebSci '17, pages 229--233, New York, NY, USA. Association for Computing
  Machinery.

\bibitem[\protect\citename{Grave \bgroup et al.\egroup }2018]{Grave:2018}
Grave, E., Bojanowski, P., Gupta, P., Joulin, A., and Mikolov, T.
\newblock (2018).
\newblock Learning word vectors for 157 languages.
\newblock In {\em Proceedings of the Eleventh International Conference on
  Language Resources and Evaluation ({LREC} 2018)}, pages 3483--3487, Miyazaki,
  Japan. European Language Resources Association (ELRA).

\bibitem[\protect\citename{Gr\"{o}ndahl \bgroup et al.\egroup
  }2018]{Grondahl:2018}
Gr\"{o}ndahl, T., Pajola, L., Juuti, M., Conti, M., and Asokan, N.
\newblock (2018).
\newblock All you need is ``{L}ove": Evading hate speech detection.
\newblock In {\em Proceedings of the 11th ACM Workshop on Artificial
  Intelligence and Security}, AISec '18, pages 2--12, New York, NY, USA.
  Association for Computing Machinery.

\bibitem[\protect\citename{Hochreiter and Schmidhuber}1997]{Hochreiter:1997}
Hochreiter, S. and Schmidhuber, J.
\newblock (1997).
\newblock Long short-term memory.
\newblock {\em Neural Computation}, 9(8):1735--1780.

\bibitem[\protect\citename{Hu \bgroup et al.\egroup }2020]{Hu:2020}
Hu, J., Ruder, S., Siddhant, A., Neubig, G., Firat, O., and Johnson, M.
\newblock (2020).
\newblock {XTREME}: A massively multilingual multi-task benchmark for
  evaluating cross-lingual generalization.
\newblock In {\em Proceedings of the 37th International Conference on Machine
  Learning}, volume 119, pages 4411--4421. PMLR.

\bibitem[\protect\citename{Jiang \bgroup et al.\egroup }2020]{Jiang:2020}
Jiang, Z., Yu, W., Zhou, D., Chen, Y., Feng, J., and Yan, S.
\newblock (2020).
\newblock Conv{BERT}: Improving {BERT} with span-based dynamic convolution.
\newblock In {\em Advances in Neural Information Processing}, volume~33, pages
  12837--12848. Curran Associates, Inc.

\bibitem[\protect\citename{Karan and {\v{S}}najder}2018]{Karan:2018}
Karan, M. and {\v{S}}najder, J.
\newblock (2018).
\newblock Cross-domain detection of abusive language online.
\newblock In {\em Proceedings of the 2nd Workshop on Abusive Language Online
  ({ALW}2)}, pages 132--137, Brussels, Belgium. Association for Computational
  Linguistics.

\bibitem[\protect\citename{Kim}2014]{Kim:2014}
Kim, Y.
\newblock (2014).
\newblock Convolutional neural networks for sentence classification.
\newblock In {\em Proceedings of the 2014 Conference on Empirical Methods in
  Natural Language Processing ({EMNLP})}, pages 1746--1751, Doha, Qatar.
  Association for Computational Linguistics.

\bibitem[\protect\citename{Kingma and Ba}2015]{Kingma:2015}
Kingma, D.~P. and Ba, J.
\newblock (2015).
\newblock Adam: {A} method for stochastic optimization.
\newblock In {\em 3rd International Conference on Learning Representations,
  {ICLR} 2015}, San Diego, CA, USA.

\bibitem[\protect\citename{Liu \bgroup et al.\egroup }2019a]{Liu:2019a}
Liu, P., Li, W., and Zou, L.
\newblock (2019a).
\newblock {NULI} at {S}em{E}val-2019 {T}ask 6: Transfer learning for offensive
  language detection using bidirectional transformers.
\newblock In {\em Proceedings of the 13th International Workshop on Semantic
  Evaluation}, pages 87--91, Minneapolis, Minnesota, USA. Association for
  Computational Linguistics.

\bibitem[\protect\citename{Liu \bgroup et al.\egroup }2019b]{Liu:2019b}
Liu, Y., Ott, M., Goyal, N., Du, J., Joshi, M., Chen, D., Levy, O., Lewis, M.,
  Zettlemoyer, L., and Stoyanov, V.
\newblock (2019b).
\newblock Ro{BERT}a: A robustly optimized {BERT} pretraining approach.
\newblock {\em arXiv preprint arXiv:1907.11692}.

\bibitem[\protect\citename{Markov and Daelemans}2021]{Markov:2021a}
Markov, I. and Daelemans, W.
\newblock (2021).
\newblock Improving cross-domain hate speech detection by reducing the false
  positive rate.
\newblock In {\em Proceedings of the Fourth Workshop on NLP for Internet
  Freedom: Censorship, Disinformation, and Propaganda}, pages 17--22, Online.
  Association for Computational Linguistics.

\bibitem[\protect\citename{Markov \bgroup et al.\egroup }2021]{Markov:2021b}
Markov, I., Ljube{\v{s}}i{\'c}, N., Fi{\v{s}}er, D., and Daelemans, W.
\newblock (2021).
\newblock Exploring stylometric and emotion-based features for multilingual
  cross-domain hate speech detection.
\newblock In {\em Proceedings of the Eleventh Workshop on Computational
  Approaches to Subjectivity, Sentiment and Social Media Analysis}, pages
  149--159, Online. Association for Computational Linguistics.

\bibitem[\protect\citename{Mathew \bgroup et al.\egroup }2021]{Mathew:2021}
Mathew, B., Saha, P., Yimam, S.~M., Biemann, C., Goyal, P., and Mukherjee, A.
\newblock (2021).
\newblock Hate{X}plain: A benchmark dataset for explainable hate speech
  detection.
\newblock In {\em Proceedings of the AAAI Conference on Artificial
  Intelligence}, volume~35, pages 14867--14875.

\bibitem[\protect\citename{Mou \bgroup et al.\egroup }2020]{Mou:2020}
Mou, G., Ye, P., and Lee, K.
\newblock (2020).
\newblock Swe2: Subword enriched and significant word emphasized framework for
  hate speech detection.
\newblock In {\em Proceedings of the 29th ACM International Conference on
  Information \& Knowledge Management}, CIKM '20, pages 1145--1154, New York,
  NY, USA. Association for Computing Machinery.

\bibitem[\protect\citename{Nguyen \bgroup et al.\egroup }2020]{Nguyen:2020}
Nguyen, D.~Q., Vu, T., and Nguyen, A.~T.
\newblock (2020).
\newblock {BERT}weet: {A} pre-trained language model for {E}nglish tweets.
\newblock In {\em Proceedings of the 2020 Conference on Empirical Methods in
  Natural Language Processing: System Demonstrations}, pages 9--14. Association
  for Computational Linguistics.

\bibitem[\protect\citename{Nobata \bgroup et al.\egroup }2016]{Nobata:2016}
Nobata, C., Tetreault, J., Thomas, A., Mehdad, Y., and Chang, Y.
\newblock (2016).
\newblock Abusive language detection in online user content.
\newblock In {\em Proceedings of the 25th International Conference on World
  Wide Web}, pages 145--153.

\bibitem[\protect\citename{Nozza}2021]{Nozza:2021}
Nozza, D.
\newblock (2021).
\newblock Exposing the limits of zero-shot cross-lingual hate speech detection.
\newblock In {\em Proceedings of the 59th Annual Meeting of the Association for
  Computational Linguistics and the 11th International Joint Conference on
  Natural Language Processing (Volume 2: Short Papers)}, pages 907--914,
  Online. Association for Computational Linguistics.

\bibitem[\protect\citename{Ousidhoum \bgroup et al.\egroup
  }2019]{Ousidhoum:2019}
Ousidhoum, N., Lin, Z., Zhang, H., Song, Y., and Yeung, D.-Y.
\newblock (2019).
\newblock Multilingual and multi-aspect hate speech analysis.
\newblock In {\em Proceedings of the 2019 Conference on Empirical Methods in
  Natural Language Processing and the 9th International Joint Conference on
  Natural Language Processing (EMNLP-IJCNLP)}, pages 4675--4684, Hong Kong,
  China. Association for Computational Linguistics.

\bibitem[\protect\citename{Pamungkas and Patti}2019]{Pamungkas:2019}
Pamungkas, E.~W. and Patti, V.
\newblock (2019).
\newblock Cross-domain and cross-lingual abusive language detection: A hybrid
  approach with deep learning and a multilingual lexicon.
\newblock In {\em Proceedings of the 57th Annual Meeting of the Association for
  Computational Linguistics: Student Research Workshop}, pages 363--370,
  Florence, Italy. Association for Computational Linguistics.

\bibitem[\protect\citename{Pamungkas \bgroup et al.\egroup
  }2020]{Pamungkas:2020}
Pamungkas, E.~W., Basile, V., and Patti, V.
\newblock (2020).
\newblock Misogyny detection in {T}witter: A multilingual and cross-domain
  study.
\newblock {\em Information Processing \& Management}, 57(6):102360.

\bibitem[\protect\citename{Paszke \bgroup et al.\egroup }2019]{Paszke:2019}
Paszke, A., Gross, S., Massa, F., Lerer, A., Bradbury, J., Chanan, G., Killeen,
  T., Lin, Z., Gimelshein, N., Antiga, L., Desmaison, A., Kopf, A., Yang, E.,
  DeVito, Z., Raison, M., Tejani, A., Chilamkurthy, S., Steiner, B., Fang, L.,
  Bai, J., and Chintala, S.
\newblock (2019).
\newblock Py{T}orch: An imperative style, high-performance deep learning
  library.
\newblock In {\em Advances in Neural Information Processing Systems},
  volume~32, pages 8024--8035, Vancouver, BC, Canada. Curran Associates, Inc.

\bibitem[\protect\citename{Pavlopoulos \bgroup et al.\egroup
  }2017]{Pavlopoulos:2017}
Pavlopoulos, J., Malakasiotis, P., Bakagianni, J., and Androutsopoulos, I.
\newblock (2017).
\newblock Improved abusive comment moderation with user embeddings.
\newblock In {\em Proceedings of the 2017 {EMNLP} Workshop: Natural Language
  Processing meets Journalism}, pages 51--55, Copenhagen, Denmark. Association
  for Computational Linguistics.

\bibitem[\protect\citename{Pedregosa \bgroup et al.\egroup
  }2011]{Pedregosa:2011}
Pedregosa, F., Varoquaux, G., Gramfort, A., Michel, V., Thirion, B., Grisel,
  O., Blondel, M., Prettenhofer, P., Weiss, R., Dubourg, V., Vanderplas, J.,
  Passos, A., Cournapeau, D., Brucher, M., Perrot, M., and Duchesnay, E.
\newblock (2011).
\newblock Scikit-learn: Machine learning in {P}ython.
\newblock {\em Journal of Machine Learning Research}, 12:2825--2830.

\bibitem[\protect\citename{Pennington \bgroup et al.\egroup
  }2014]{Pennington:2014}
Pennington, J., Socher, R., and Manning, C.~D.
\newblock (2014).
\newblock Glo{V}e: Global vectors for word representation.
\newblock In {\em Proceedings of the 2014 Conference on Empirical Methods in
  Natural Language Processing (EMNLP)}, pages 1532--1543, Doha, Qatar.
  Association for Computational Linguistics.

\bibitem[\protect\citename{Poletto \bgroup et al.\egroup }2021]{Poletto:2021}
Poletto, F., Basile, V., Sanguinetti, M., Bosco, C., and Patti, V.
\newblock (2021).
\newblock Resources and benchmark corpora for hate speech detection: {A}
  systematic review.
\newblock {\em Language Resources and Evaluation}, 55(2):477--523.

\bibitem[\protect\citename{{\c{S}}ahinu{\c{c}} and Toraman}2021]{Sahinuc:2021}
{\c{S}}ahinu{\c{c}}, F. and Toraman, C.
\newblock (2021).
\newblock Tweet length matters: A comparative analysis on topic detection in
  microblogs.
\newblock In {\em Advances in Information Retrieval}, pages 471--478, Cham.
  Springer International Publishing.

\bibitem[\protect\citename{Sanguinetti \bgroup et al.\egroup
  }2018]{Sanguinetti:2018}
Sanguinetti, M., Poletto, F., Bosco, C., Patti, V., and Stranisci, M.
\newblock (2018).
\newblock An {I}talian {T}witter corpus of hate speech against immigrants.
\newblock In {\em Proceedings of the Eleventh International Conference on
  Language Resources and Evaluation ({LREC} 2018)}, pages 2798--2805, Miyazaki,
  Japan. European Language Resources Association (ELRA).

\bibitem[\protect\citename{Schweter}2020]{Schweter:2020}
Schweter, S.
\newblock (2020).
\newblock {BERT}urk - {BERT} models for {T}urkish.
\newblock https://doi.org/10.5281/zenodo.3770924.
\newblock Accessed: 2021-10-15.

\bibitem[\protect\citename{Sharma \bgroup et al.\egroup }2018]{Sharma:2018}
Sharma, S., Agrawal, S., and Shrivastava, M.
\newblock (2018).
\newblock Degree based classification of harmful speech using {T}witter data.
\newblock In {\em Proceedings of the First Workshop on Trolling, Aggression and
  Cyberbullying ({TRAC}-2018)}, pages 106--112, Santa Fe, New Mexico, USA.
  Association for Computational Linguistics.

\bibitem[\protect\citename{Shoeybi \bgroup et al.\egroup }2019]{Shoeybi:2019}
Shoeybi, M., Patwary, M., Puri, R., LeGresley, P., Casper, J., and Catanzaro,
  B.
\newblock (2019).
\newblock Megatron-{LM}: Training multi-billion parameter language models using
  model parallelism.
\newblock {\em arXiv preprint arXiv:1909.08053}.

\bibitem[\protect\citename{Sood \bgroup et al.\egroup }2012]{Sood:2012}
Sood, S., Antin, J., and Churchill, E.
\newblock (2012).
\newblock Profanity use in online communities.
\newblock In {\em Proceedings of the SIGCHI Conference on Human Factors in
  Computing Systems}, pages 1481--1490.

\bibitem[\protect\citename{Swamy \bgroup et al.\egroup }2019]{Swamy:2019}
Swamy, S.~D., Jamatia, A., and Gamb{\"a}ck, B.
\newblock (2019).
\newblock Studying generalisability across abusive language detection datasets.
\newblock In {\em Proceedings of the 23rd Conference on Computational Natural
  Language Learning (CoNLL)}, pages 940--950, Hong Kong, China. Association for
  Computational Linguistics.

\bibitem[\protect\citename{Turc \bgroup et al.\egroup }2021]{Turc:2021}
Turc, I., Lee, K., Eisenstein, J., Chang, M.-W., and Toutanova, K.
\newblock (2021).
\newblock Revisiting the primacy of {E}nglish in zero-shot cross-lingual
  transfer.
\newblock {\em arXiv preprint arXiv:2106.16171}.

\bibitem[\protect\citename{Twitter}2021]{TwitterReport:2021}
Twitter.
\newblock (2021).
\newblock {Twitter Transparency Report}.
\newblock
  https://transparency.twitter.com/en/reports/rules-enforcement.html\#2020-jul-dec.
\newblock Accessed: 2021-10-15.

\bibitem[\protect\citename{Unsv{\aa}g and Gamb{\"a}ck}2018]{Unsvaag:2018}
Unsv{\aa}g, E.~F. and Gamb{\"a}ck, B.
\newblock (2018).
\newblock The effects of user features on {T}witter hate speech detection.
\newblock In {\em Proceedings of the 2nd Workshop on Abusive Language Online
  (ALW2)}, pages 75--85.

\bibitem[\protect\citename{Vaswani \bgroup et al.\egroup }2017]{Vaswani:2017}
Vaswani, A., Shazeer, N., Parmar, N., Uszkoreit, J., Jones, L., Gomez, A.~N.,
  Kaiser, {\L}., and Polosukhin, I.
\newblock (2017).
\newblock Attention is all you need.
\newblock In {\em Advances in Neural Information Processing Systems},
  volume~30, pages 5998--6008. Curran Associates, Inc.

\bibitem[\protect\citename{Waseem}2016]{Waseem:2016}
Waseem, Z.
\newblock (2016).
\newblock Are you a racist or am {I} seeing things? {A}nnotator influence on
  hate speech detection on {T}witter.
\newblock In {\em Proceedings of the First Workshop on NLP and Computational
  Social Science}, pages 138--142.

\bibitem[\protect\citename{Wolf \bgroup et al.\egroup }2020]{Wolf:2019}
Wolf, T., Debut, L., Sanh, V., Chaumond, J., Delangue, C., Moi, A., Cistac, P.,
  Rault, T., Louf, R., Funtowicz, M., Davison, J., Shleifer, S., von Platen,
  P., Ma, C., Jernite, Y., Plu, J., Xu, C., Le~Scao, T., Gugger, S., Drame, M.,
  Lhoest, Q., and Rush, A.
\newblock (2020).
\newblock Transformers: State-of-the-art natural language processing.
\newblock In {\em Proceedings of the 2020 Conference on Empirical Methods in
  Natural Language Processing: System Demonstrations}, pages 38--45, Online.
  Association for Computational Linguistics.

\bibitem[\protect\citename{Zampieri \bgroup et al.\egroup }2020]{Zampieri:2020}
Zampieri, M., Nakov, P., Rosenthal, S., Atanasova, P., Karadzhov, G., Mubarak,
  H., Derczynski, L., Pitenis, Z., and {\c{C}}{\"o}ltekin, {\c{C}}.
\newblock (2020).
\newblock {S}em{E}val-2020 {T}ask 12: Multilingual offensive language
  identification in social media ({O}ffens{E}val 2020).
\newblock In {\em Proceedings of the Fourteenth Workshop on Semantic
  Evaluation}, pages 1425--1447, Barcelona (Online). International Committee
  for Computational Linguistics.

\bibitem[\protect\citename{Zimmerman \bgroup et al.\egroup
  }2018]{Zimmerman:2018}
Zimmerman, S., Kruschwitz, U., and Fox, C.
\newblock (2018).
\newblock Improving hate speech detection with deep learning ensembles.
\newblock In {\em Proceedings of the Eleventh International Conference on
  Language Resources and Evaluation (LREC 2018)}, pages 2546--2553, Miyazaki,
  Japan. European Language Resources Association (ELRA).

\end{thebibliography}

% \section{Language Resource References}
% \label{lr:ref}
% \bibliographystylelanguageresource{lrec2022-bib}
% \bibliographylanguageresource{languageresource}

\end{document}